%% file: QuSecNets.tex
\documentclass[10pt, conference, letterpaper]{IEEEtran}
\IEEEoverridecommandlockouts

\usepackage{setspace}

\usepackage{booktabs} 
\usepackage{cite}
\usepackage{amsmath}
\usepackage{amssymb}
\usepackage{amsfonts}
\usepackage{float}
\usepackage{graphicx}
\usepackage{xcolor}
\usepackage{colortbl}
\usepackage{multirow}
\usepackage{mathtools}
\usepackage{algorithm}
\usepackage{algorithmicx}
\usepackage{algpseudocode}
\usepackage{tabto}
\usepackage{todonotes}
\usepackage{hhline}
\usepackage{enumitem}
\usepackage[belowskip=-8pt,aboveskip=8pt,labelfont={it,bf},font=small]{caption}
\def\BibTeX{{\rm B\kern-.05em{\sc i\kern-.025em b}\kern-.08em
    T\kern-.1667em\lower.7ex\hbox{E}\kern-.125emX}}
    
    \makeatletter
\renewcommand\footnoterule{%
  \kern-3\p@
  \hrule\@width 0.5\columnwidth
  \kern2.6\p@}
  \makeatother
\DeclareRobustCommand*{\IEEEauthorrefmark}[1]{%
  \raisebox{0pt}[0pt][0pt]{\textsuperscript{\footnotesize #1}}%
}

\usepackage{fancyhdr,lipsum}
\fancypagestyle{firstpage}{
  \fancyhf{}
  \fancyhead[L]{Published as a conference paper at IOLTS 2019}
}
\begin{document}
\title{QuSecNets: Quantization-based Defense Mechanism for Securing Deep Neural Network against Adversarial Attacks }

\author{\IEEEauthorblockN{
    Faiq Khalid\IEEEauthorrefmark{1}\IEEEauthorrefmark{,}\IEEEauthorrefmark{*}\thanks{*Faiq Khalid and Hassan Ali both are lead authors, and have equal contributions.},
    Hassan Ali\IEEEauthorrefmark{2}\IEEEauthorrefmark{,}\IEEEauthorrefmark{*},
    Hammad Tariq\IEEEauthorrefmark{2},
    Muhammad Abdullah Hanif\IEEEauthorrefmark{1}, \\ 
    Semeen Rehman\IEEEauthorrefmark{1}, 
    Rehan Ahmed\IEEEauthorrefmark{2}, 
    Muhammad Shafique\IEEEauthorrefmark{1}}
    \IEEEauthorblockA{\IEEEauthorrefmark{1} Technische Universit\"at Wien (TU Wien), Vienna, Austria\\
    Email: \{faiq.khalid, muhammad.hanif, semeen.rehman,  muhammad.shafique\}@tuwien.ac.at}
    \IEEEauthorblockA{\IEEEauthorrefmark{2} National University of Sciences and Technology (NUST), Islamabad, Pakistan\\
    Email: \{rehan.ahmed, hali.msee17, htariq.msee17\}@seecs.edu.pk} \vspace{-25pt}%
    }

\maketitle
\thispagestyle{firstpage}
\begin{abstract}
	Adversarial examples have emerged as a significant threat to machine learning algorithms, especially to the convolutional neural networks (CNNs). In this paper, we propose two quantization-based defense mechanisms, Constant Quantization (CQ) and Trainable Quantization (TQ), to increase the robustness of CNNs against adversarial examples. CQ quantizes input pixel intensities based on a ``fixed'' number of quantization levels, while in TQ, the quantization levels are ``iteratively learned during the training phase'', thereby providing a stronger defense mechanism. We apply the proposed techniques on undefended CNNs against different state-of-the-art adversarial attacks from the open-source \textit{Cleverhans} library. The experimental results demonstrate 50\%-96\% and 10\%-50\% increase in the classification accuracy of the perturbed images generated from the MNIST and the CIFAR-10 datasets, respectively, on commonly used CNN (Conv2D(64, 8x8) - Conv2D(128, 6x6) - Conv2D(128, 5x5) - Dense(10) - Softmax()) available in \textit{Cleverhans} library. \\      
\end{abstract}

\input{Sections/Sec1_Introduction.tex}
\input{Sections/Sec2_SoA.tex}
\input{Sections/Sec3_QuSecNets.tex}
\input{Sections/Sec4_Results.tex}
\input{Sections/Sec5_Comparison.tex}
\input{Sections/Sec6_Conclusion.tex}
\input{Sections/Sec_ack.tex}
\bibliographystyle{IEEEtran.bst}
\bibliography{QuSecNets.bbl}

\end{document}

%% file: Sections/Sec1_Introduction.tex
\section{Introduction}\label{introduction}
Over the past few years, machine learning algorithms, especially, convolutional neural networks (CNNs) have emerged as a prime solution for complex classification and recognition applications in many safety-critical domains, e.g., autonomous driving and smart healthcare \cite{stilgoe2018machine}\cite{ratasich2019roadmap}\cite{hailesellasie2018fpga}. However, due to their data dependencies, especially for complex classification tasks, CNNs are inherently vulnerable to several security threats, e.g., data poisoning \cite{DBLP:conf/date/0001TBHKHR18,DBLP:journals/corr/abs-1811-01031,khalid2018fademl2,khalid2018security,hanif2018robust, kriebel2018robustness}, model stealing \cite{duddu2018stealing}\cite{tramer2016stealing} and adversarial attacks \cite{yuan2019adversarial}\cite{khalid2019red}.  

Adversarial attacks during the inference have emerged as one of the most powerful attacks on CNNs because of their imperceptibility to subjective evaluation, and correlation and structural similarity analysis \cite{DBLP:journals/corr/abs-1811-01031}. These attacks do not change the CNN model (structure and parameters) rather exploit their data dependent behavior using back-propagation and gradient descent to perform misclassification and confidence reduction attacks \cite{goodfellow2014explaining}. \textit{Therefore, there is a dire need of developing defense mechanisms to protect the CNN inference.} 

Several defense mechanisms against adversarial attacks have been proposed. The most notable are adversarial training \cite{DBLP:journals/corr/SzegedyZSBEGF13,DBLP:journals/corr/ShenJGZ17,goodfellow2014explaining} and gradient masking \cite{DBLP:conf/sp/PapernotM0JS16}\cite{DBLP:journals/corr/abs-1807-06714}. These defense mechanisms either change the DNN structure, modify the training procedure or train it against only known adversarial attacks, which limit their defense scope to known vulnerabilities only. Several counter-attacks have been proposed to compromise these defense mechanisms \cite{DBLP:journals/corr/CarliniW16a,DBLP:journals/corr/CarliniW16,DBLP:journals/corr/abs-1711-08478}. Researchers have also exploited the preprocessing filtering to develop defense mechanisms against adversarial examples. However, these defense mechanisms can be compromised by modifying the gradient estimation algorithm in such a way that it can incorporate the effects of the preprocessing filters into their respective optimization algorithms, especially in black-box settings \cite{khalid2018fademl2}. Similarly, preprocessing quantization is exploited to reduce the effect of adversarial noise \cite{DBLP:conf/ndss/Xu0Q18}\cite{DBLP:journals/corr/abs-1807-06714} because of its inherent property of being insensitive to small perturbations. However, the key challenge in such techniques is to \textit{define and learn appropriate quantization levels which ensure high perceptibility of the attack noise and do not affect the classification accuracy in both the attack-free and the attack scenario.}   
\begin{figure*}[!t]
	\centering
	\includegraphics[width=1\linewidth]{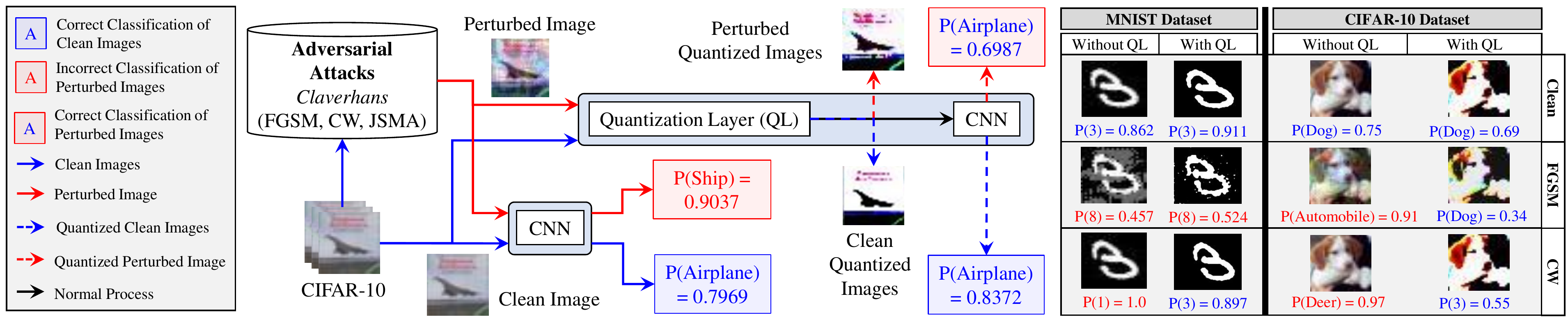} 
	\caption{\textit{Experimental setup for analyzing the effects of quantization on the classification of clean and perturbed images (generated using state-of-the-art adversarial attacks, i.e., the FGSM and the C\&W, in the open-source \textit{Cleverhans} library \cite{DBLP:journals/corr/GoodfellowPM16}). P(x) represents the probability with which the CNN classifies a particular input image to a class "x".}}
	\vskip -0.1in
	\label{fig:MA}
\end{figure*} 
\subsection{Motivational Analysis} \label{subsec:MA}
To study and compare the effects of constant quantization on the classification of clean samples versus perturbed samples, we performed the state-of-the-art adversarial attacks from the \textit{Cleverhans} \cite{DBLP:journals/corr/GoodfellowPM16} library, i.e., Fast Gradient Sign Method (FGSM) and Carlini and Wagner (CW), on one of the commonly used CNN structure\footnote{Conv(64, 8x8) - Conv(128, 6x6) - Conv(128, 5x5) - Dense(10) - Softmax()} from the \textit{Cleverhans} library for the MNIST and the CIFAR-10 datasets. For this analysis, we consider the CNN both \textit{with} and \textit{without} a quantization layer at the input of the CNN, as shown in Fig.~\ref{fig:MA}. Our experimental analysis shows that by quantizing the input of the CNN, \textit{the confidence with which a particular clean image is correctly classified remains almost the same, while the confidence with which the perturbed image is incorrectly classified decreases}. For example, in case of clean images, the confidence with which the CNN correctly classifies the images from the MINST (``3'') and the CIFAR-10 (Dog) remains almost in similar range, i.e., 0.862 to 0.911 (for ``3'') and 0.75 to 0.69 (for Dog), as shown in Fig.~\ref{fig:MA}. In case of perturbed images, the CNN without input quantization incorrectly classifies the image of an airplane (from the CIFAR-10 dataset) as a ship with the confidence 0.9037, however, the same perturbed image when passed through the CNN with input quantization is correctly classified as an airplane with the confidence 0.6987. On the other hand, the confidence with which the CNN with input quantization incorrectly classify the FGSM-based perturbed image of label ``3" as ``8" increases from 0.457 to 0.524. These results highlight the \textit{unpredictable behavior in constant quantization-based defenses}. Therefore, to address this research challenge, a more flexible and comprehensive defense strategy is required which can make the adversarial noise perceptible or decrease the miclassification probability under the attack scenarios.  

\subsection{Novel Contribution}
To address the above-mentioned research challenges, we propose a \textit{trainable} quantization-based defense mechanism (QuSecNets) against adversarial attacks. Our contributions are:.  
\begin{enumerate}[leftmargin=*]
    \item \textbf{Quantization-based Defense (Section III):} We study quantization as a defense against adversarial attacks by introducing and learning an additional quantization layer at the input of the CNNs.
    
    \item \textbf{Trainable Quantization (Section III)}: To address the \textit{unpredictable behavior} of quantization (as highlighted in the motivational analysis and Fig.~\ref{fig:MA}) in image classification, we propose a \textit{trainable quantization-based defense} against adversarial attacks, which learns the thresholds (i.e., quantization levels) during the training phase.
\end{enumerate}
To illustrate the effectiveness of \textit{QuSecNets} with Constant Quantization (CQ) and Trainable Quantization (TQ), we evaluate it against different adversarial attacks from the open-source \textit{Cleverhans} library (i.e., the FGSM, the C\&W and the Jacobian  Saliency  Map  Attack  (JSMA)) on an undefended CNN\footnote{Conv(64, 8x8) - Conv(128, 6x6) - Conv(128, 5x5) - Dense(10) - Softmax()} from the \textit{Cleverhans} available in the \textit{Cleverhans} library) for the MNIST and the CIFAR-10 datasets (Section \ref{results}). We also compare it with the state-of-the-art defenses (Section \ref{comparison}). The experimental results show that \textit{QuSecNets} demonstrates 50\%-96\% and 10\%-50\% increase in the classification accuracy of the perturbed images generated from the MNIST and the CIFAR-10 datasets, respectively.


%% file: Sections/Sec2_SoA.tex
\section{State-of-the-Art Defenses against Adversarial Attacks on DNNs} \label{SoA}
This section provides a brief overview of different adversarial attacks and the state-of-the-art defense mechanisms against them.

\subsection{Adversarial Attacks}
Adversarial attacks are imperceptible perturbations in the input image by an adversary who aims at manipulating a DNN for a targeted or an un-targeted misclassification. The strength/robustness\footnote{To achieve the maximum confidence for misclassification with minimum perturbations} and imperceptibility against subjective evaluations of these attacks are highly dependent on their optimization algorithms. Based on the attack strategies and optimization algorithms, these adversarial attacks can be categorized as follows \cite{DBLP:journals/corr/RauberBB17}:
\begin{enumerate}[leftmargin=*]
    \item \textit{Gradient-based attacks} generate adversarial noise based on the gradient of the loss function with respect to the inputs, e.g., the FGSM \cite{goodfellow2014explaining}, the JSMA \cite{DBLP:journals/corr/PapernotMJFCS15}, the Basic Iterative Method (BIM) \cite{DBLP:journals/corr/KurakinGB16}, the C\&W \cite{DBLP:journals/corr/CarliniW16a}, and the DeepFool \cite{DBLP:conf/cvpr/Moosavi-Dezfooli16}.
    
    \item \textit{Decision-based attacks} do not require gradient estimation, however, they utilize the output decision of the CNN to compute adversarial noise \cite{khalid2019red}, e.g., the Point-wise Attack  \cite{DBLP:journals/corr/RauberBB17}\cite{dong2019efficient}\cite{liu2019geometry} and the Additive Gaussian Noise Attack \cite{DBLP:journals/corr/RauberBB17}.
    
    \item \textit{Score-based attacks} analyze the statistical (or probabilistic) behavior of individual input components to estimate the corresponding gradients with respect to loss function, e.g., the Single-Pixel Attack \cite{DBLP:journals/corr/NarodytskaK16} and the Local Search Attack \cite{DBLP:journals/corr/NarodytskaK16}.
    
\end{enumerate}
\subsection{Defense Mechanisms} \label{subsec:defense}
To improve and ensure the security of ML-based applications, several defense strategies have been proposed based on DNN masking \cite{DBLP:conf/ndss/Xu0Q18}, gradient masking \cite{DBLP:conf/sp/PapernotM0JS16}, training for known adversarial attacks \cite{DBLP:journals/corr/SzegedyZSBEGF13} and preprocessing of the inputs \cite{DBLP:conf/ndss/Xu0Q18}\cite{DBLP:journals/corr/abs-1807-06714}\cite{DBLP:journals/corr/ShenJGZ17}\cite{DBLP:journals/corr/abs-1711-08478}\cite{DBLP:journals/corr/abs-1805-06605}. Based on these defense strategies, the defenses can be classified as follows: 
\begin{enumerate}[leftmargin=*]
    \item \textit{Adversarial Learning} is one of the most commonly used approach which trains the CNNs for known adversarial attacks \cite{goodfellow2014explaining}.
    
    \item \textit{Gradient Masking-based Defenses} either mask the gradients or the whole DNN. For example, defensive distillation masks the gradients of the network, but it is only valid for gradient-based attacks \cite{liu2018feature,DBLP:journals/corr/abs-1805-06605,buckman2018thermometer,DBLP:journals/corr/abs-1803-01442,DBLP:journals/corr/abs-1711-01991}. However, it can be neutralize by empirically inferring the gradients using different loss functions \cite{DBLP:journals/corr/CarliniW16}.
    
    \item {Pre-processing} has emerged as one of the prime defense mechanisms against adversarial attacks \cite{khalid2018fademl2}\cite{rakin2018blind}. For example, feature squeezing technique uses binary quantization \cite{DBLP:conf/ndss/Xu0Q18}.
    
    \item MagNet is another preprocessing-based defense \cite{DBLP:conf/ccs/MengC17} which trains one or more detector networks and a reformer network to defend against adversarial attacks. The detector networks are responsible for detecting the adversarial examples. If the input is detected as clean, the reformer network generates a variant of input and feeds to the DNN. However, it cannot counter the C\&W attack \cite{DBLP:journals/corr/abs-1711-08478} and it is also computationally expensive. 
    
    \item \textit{Generative Adversarial Network (GAN)-based defenses} has been proposed \cite{DBLP:journals/corr/ShenJGZ17} that use a generator network, which generates adversarial examples, to train the CNN for known adversarial attacks. However, this strategy can be countered by adversarial attacks which purposely introduce noise to fool the generator \cite{DBLP:journals/corr/abs-1711-08478}. In addition, these defenses are computationally very expensive and often impractical, especially in resource-constrained applications.
    
    \item Another effective strategy used for increasing the robustness of neural networks is to apply \textit{data augmentation} techniques during the training \cite{DBLP:conf/ccs/ZantedeschiNR17}. This strategy is somewhat similar to the adversarial training and hence, possesses the same limitation, i.e., the network becomes robust to only known adversarial attacks.
    
\end{enumerate}

%% file: Sections/Sec3_QuSecNets.tex
\section{Quantization-based Defense for CNNs} \label{QusecNets}
\begin{figure}[!t]
	\centering
	\includegraphics[width=1\linewidth]{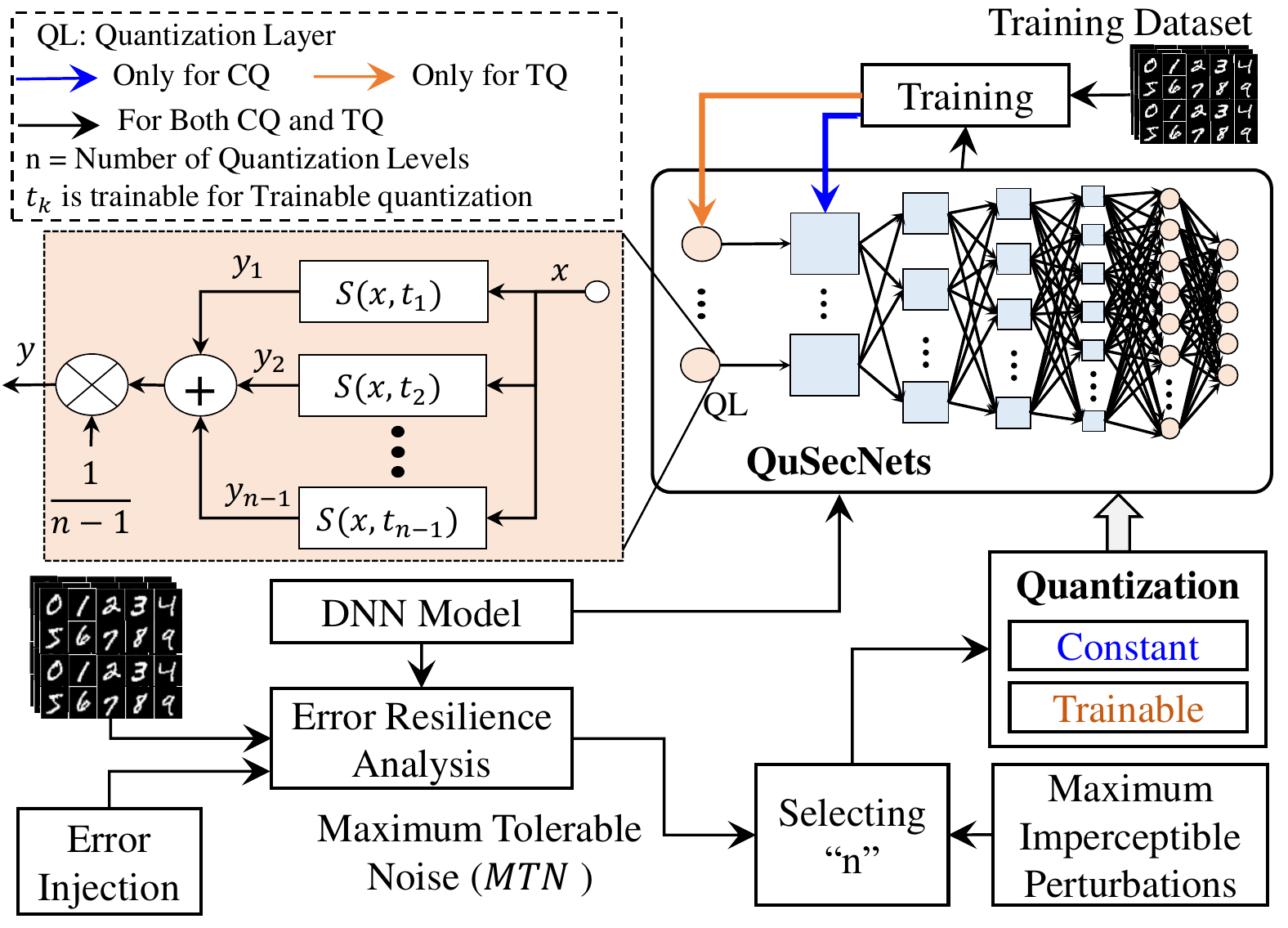} 
	\caption{\textit{Quantization-based defense mechanism, QuSecNets, that integrates constant or trainable quantization layer to reduce the effects of adversarial noise. In constant quantization the number of quantization levels and the corresponding threshold values remain unchanged while training, and are computed based on the error resilience of an application and maximum imperceptible perturbations. However, the trainable quantization defines the quantization levels during the training.}}
	\vskip -0.1in
	\label{fig:QuSecNets}
\end{figure}

To address the above-mentioned limitations of the state-of-the-art defenses (Section \ref{subsec:defense}), we propose to leverage the quantization with constant quantization levels and trainable quantization levels (to address the uncertainty issue, as discussed in Section \ref{subsec:MA}) to develop a flexible defense mechanism. However, this requires addressing the following research challenges:
\begin{enumerate}[leftmargin=*]
    \item How to identify the appropriate number of quantization levels to achieve desired classification accuracy?
    \item How to identify the appropriate values of quantization levels to increase the perceptibility of adversarial noise while maintaining the classification accuracy? 
\end{enumerate}
\textbf{QuSecNets: A Quantization-based Defense Mechanism:} To address these research challenges, we propose a quantization-based defense mechanism, QuSecNets, which consists of the following steps, also shown in Fig.~\ref{fig:QuSecNets}:
\begin{enumerate}[leftmargin=*]
    \item First, the number of quantization levels are selected based on error resilience of an applications and the maximum imperceptible perturbations\footnote{The maximum perturbations that can be added without making the attack noise perceptible.}, as shown in Fig.~\ref{fig:QuSecNets}. In error resilience analysis, the Maximum Tolerable Noise (MTN) is estimated by performing error injection with varying noise strengths (e.g., using the methodology of ~\cite{hanif2018error}). By analyzing the $MTN$ and $epsilon$, an appropriate  number of quantization levels $n$ can be selected.  
    \textit{Note, in the case of constant quantization, the quantization levels are linearly distributed.} 
    
    \item Then we integrate the quantization layer at the input of the CNN. The quantization layer has one-to-one relation with the input layer, i.e., for each input pixel there is a separate quantization function. This function is defined as the average value of $``n-1"$ sigmoid functions $``S(x,\ t_k)"$ (see Fig.~\ref{fig:QuSecNets}) and can be formulated as:
    \begin{equation}
    \footnotesize
      y = \frac{1}{n-1} \times \sum_{k=1}^{n-1} S(x,t_k) = \frac{1}{n-1} \times \sum_{k=1}^{n-1} \Bigg(\frac{1}{1+e^{-z(x-t_k)}}\Bigg)
    \label{Eq:eq1}
    \end{equation}
    Where $x$, $z$ and $t_k$ are the intensity of a single pixel from the input image, the scalar constant and the threshold values (i.e., quantization levels) which \textit{may} (in the case of trainable quantization) or \textit{may not} (for constant quantization) be trained, respectively. Moreover, $n$ represents the number of quantization levels, e.g., if the number of quantization levels is 2 (i.e., one sigmoid function), all the values below the threshold $t$ are pushed towards 0, while those above $t$ are pushed towards the maximum value, which is 1 in our case. \textit{Note, we chose the sigmoid function to model the quantization because it is differentiable, thereby trainable using back-propagation algorithm.}  
    
    \item After integrating the quantization layer in the CNN, we train the modified network, depending upon the quantization methodology. For example, in the case of constant quantization, we use predefined quantization levels which are linearly distributed in our case. However, in the case of trainable quantization, we propose to leverage the back-propagation algorithm to compute the appropriate threshold values based on an optimization function.
    
\end{enumerate}
In \textbf{trainable quantization}, one of the most important design challenges is \textit{to adjust the thresholds $(t_k)$ based on the output prediction/classification probabilities of the CNNs}. To address this challenge, we define a cost function (similar to the traditional cost function used for CNNs training) which computes the difference between the actual and targeted predictions. 
    \begin{equation}
    \footnotesize
        cost = \frac{1}{c} \times \sum_{c=1}^{C} (P(x=c) - P'(x=c))^2
        \label{Eq:eq2}
    \end{equation}
Where, $P'(x=c)$ and $P(x=c)$ represent the probability of input $x$ being classified as class $c$ and the ground truth probability of input $x$ being classified as class $c$, respectively. We optimize this cost function using back-propagation algorithm, which is commonly used in CNN training, with the help of the following equations.
    \begin{equation}
    \footnotesize
        t_{k,new} =  t_{k,old} - \eta \frac{\partial (cost)}{\partial (t_{k,old})}
        \label{Eq:eq3}
    \end{equation}
Where, $\eta$ represents the learning rate of the back-propagation algorithm. 
The effect of threshold on cost function, i.e.,  $\frac{\partial (cost)}{\partial (t_{k,old}}$, can be computed by applying the chain rule:   
    \begin{equation}
    \footnotesize
         \frac{\partial (cost)}{\partial (t_{k,old})} = \frac{\partial (cost)}{\partial y} \times \frac{\partial y}{\partial (t_{k,old})}
        \label{Eq:eq4}
    \end{equation}
    
    \begin{equation}
    \footnotesize
         \frac{\partial y}{\partial (t_{k,old})} =  \Bigg(\frac{-1}{n-1} \times \sum^{n-1}_{k=1}  z \times y_k \times (1-y_k)\Bigg)
        \label{Eq:eq5}
    \end{equation}
    
    \begin{equation}
    \footnotesize
        \frac{\partial (cost)}{\partial y} = \sum_{j} (w_{jk} \times \delta_j^{(l+1)}) = \delta_k^{(l)}
        \label{Eq:eq6}
    \end{equation}

Where $\delta_j^{(l+1)}$ represents the sensitivity of the $j^{th}$ neuron in the $(l+1)^{th}$ layer, while $w_{jk}$ is the weight connecting the $k^{th}$ neuron in $l^{th}$ layer  to the $j^{th}$ neuron in $(l+1)^{th}$ layer. 

Using Equations \ref{Eq:eq4}, \ref{Eq:eq5} and \ref{Eq:eq6}, Equation \ref{Eq:eq3} can be re-written as:
    \begin{equation}
    \footnotesize
        t_{1,new} =  t_{1,old} - \eta \times \frac{d}{dy} cost \times \frac{z \times(-y(1-y))}{n-1}
        \label{Eq:eq10}
    \end{equation}
    \begin{equation}
    \footnotesize
        Update\ factor =  \frac{d}{dy} cost \times \frac{z \times(-y(1-y))}{n-1}
        \label{Eq:eq10}
    \end{equation}

To control the update factor, the value of scalar constant $z$ should not be very high or very low. So, based on our empirical analysis, we set the range of $z$ between 5 and 50.  


%% file: Sections/Sec4_Results.tex
\section{Results and Discussion} \label{results}
\begin{figure*}[!t]
	\centering
	\includegraphics[width=1\linewidth]{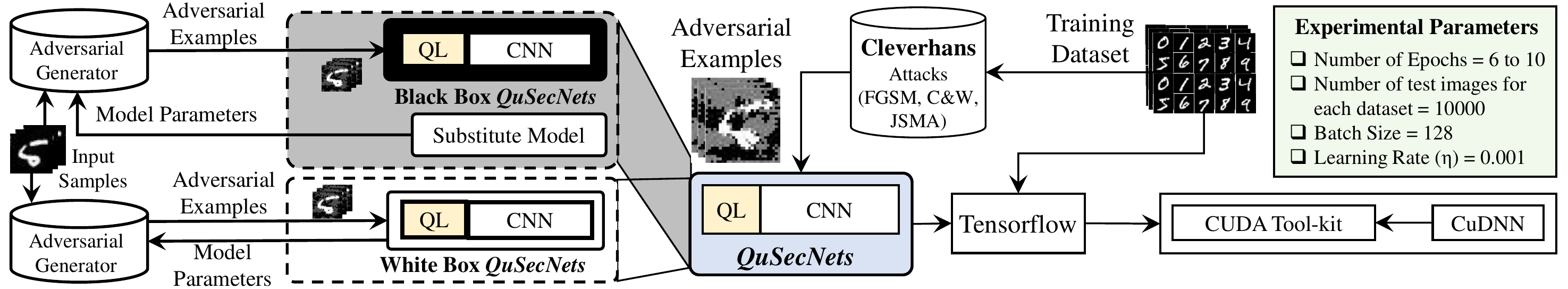} 
	\caption{\textit{Experimental setup for evaluating the proposed \textit{QuSecNets} in both black-box and white-box settings for the commonly used CNN from the \textit{Cleverhans} library against the state-of-the-art adversarial attacks (also avaialable in the \textit{Cleverhans} library), i.e., FGSM, C\&W and JSMA.}}
	\vskip -0.1in
	\label{fig:Exp_setup}
\end{figure*}

To illustrate the effectiveness of the proposed \textit{QuSecNets}, we integrate the quantization layer with one of the commonly used CNNs from the open-source \textit{Cleverhans} library and evaluate it against the state-of-the-art adversarial attacks, i.e., the FGSM, the C\&W, and the JSMA (also available in the open-source \textit{Cleverhans} library). 

\subsection{Experimental Setup}
\begin{figure}[!t]
	\centering
	\includegraphics[width=1\linewidth]{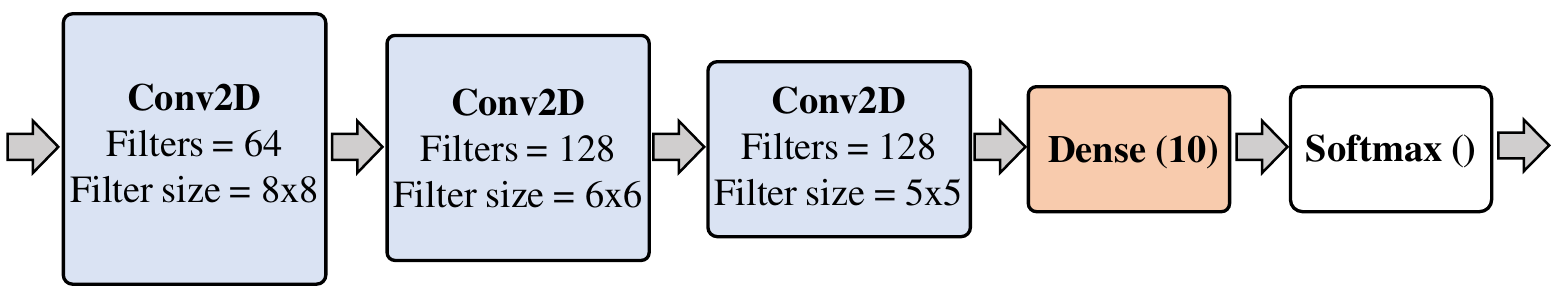}
	\caption{\textit{The CNN model used for evaluation taken from \textit{Clvehans Library}.}}
	\vskip -0.1in
	\label{fig:vggnet}
\end{figure}

We perform several experiments and analyses using the following experimental setup and settings/configurations also shown in Fig.~\ref{fig:Exp_setup}: 
\begin{enumerate}[leftmargin=*]
    \item \textbf{CNN:} We use the following CNN structure from the \textit{Cleverhans} library: Conv2D(64, 8x8) - Conv2D(128, 6x6) - Conv2D(128, 5x5) - Dense(10) - Softmax(), also shown in Fig. \ref{fig:vggnet}. 
    \item \textbf{Dataset:} We train and test the above-mentioned CNN architecture for the CIFAR-10 and the MNIST datasets.
    \item \textbf{Adversarial Attacks:} We performed state-of-the-art adversarial attacks from the \textit{Cleverhans} library, like FGSM, JSMA, and C\&W. 
    \item \textbf{Threat Models:} For a comprehensive analysis of \textit{QuSecNets}, we assume both \textit{white-box} and \textit{black-box} threat models, also shown in Fig.~\ref{fig:Exp_setup}. In both the models, an adversary can only alter the inputs, however, in the white-box scenario the adversary knows the parameters and structure of the CNN.
    \item \textbf{Quantization Settings:} We empirically identified the scalar constant value $z$ to be 50 for the constant quantization, and 5 for the trainable quantization.
\end{enumerate}
   \begin{figure}[b]
    	\centering
    	\includegraphics[width=1\linewidth]{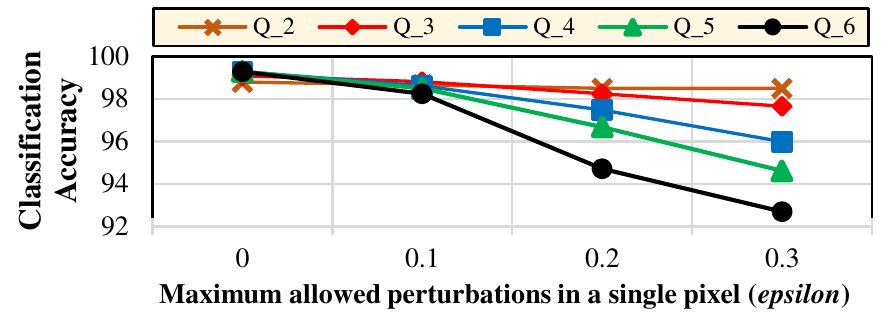}
    	\caption{\textit{The effects of varying the number of quantization levels in CQ on classification accuracy of the adversarial examples generated using the FGSM attack for different values of $epsilon$. Note, $Q\_n$ represents the quantization with $n$ number of quantization levels and $epsilon$ represents the maximum allowed perturbations in a single pixel, where the range of a pixel is from $0$ to $1$.}}
    	\label{fig:epsilon_trend}
    	\vskip -0.1in
    \end{figure}
    \begin{figure}[!t]
	\centering
	\includegraphics[width=1\linewidth]{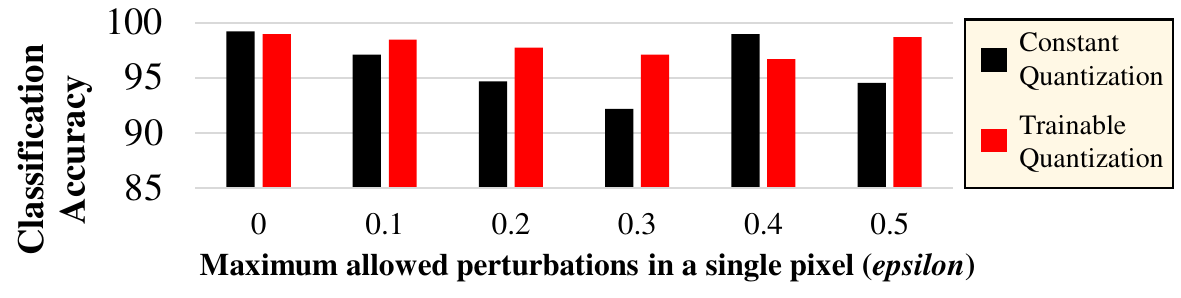} 
	\caption{\textit{Comparative analysis of CQ and TQ. The adversarial examples for this analysis are generated using the FGSM attack with MNIST.}}
	\vskip -0.1in
	\label{fig:effect_VQ}
\end{figure}
    \begin{figure*}[!t]
    	\centering
    	\includegraphics[width=1\linewidth]{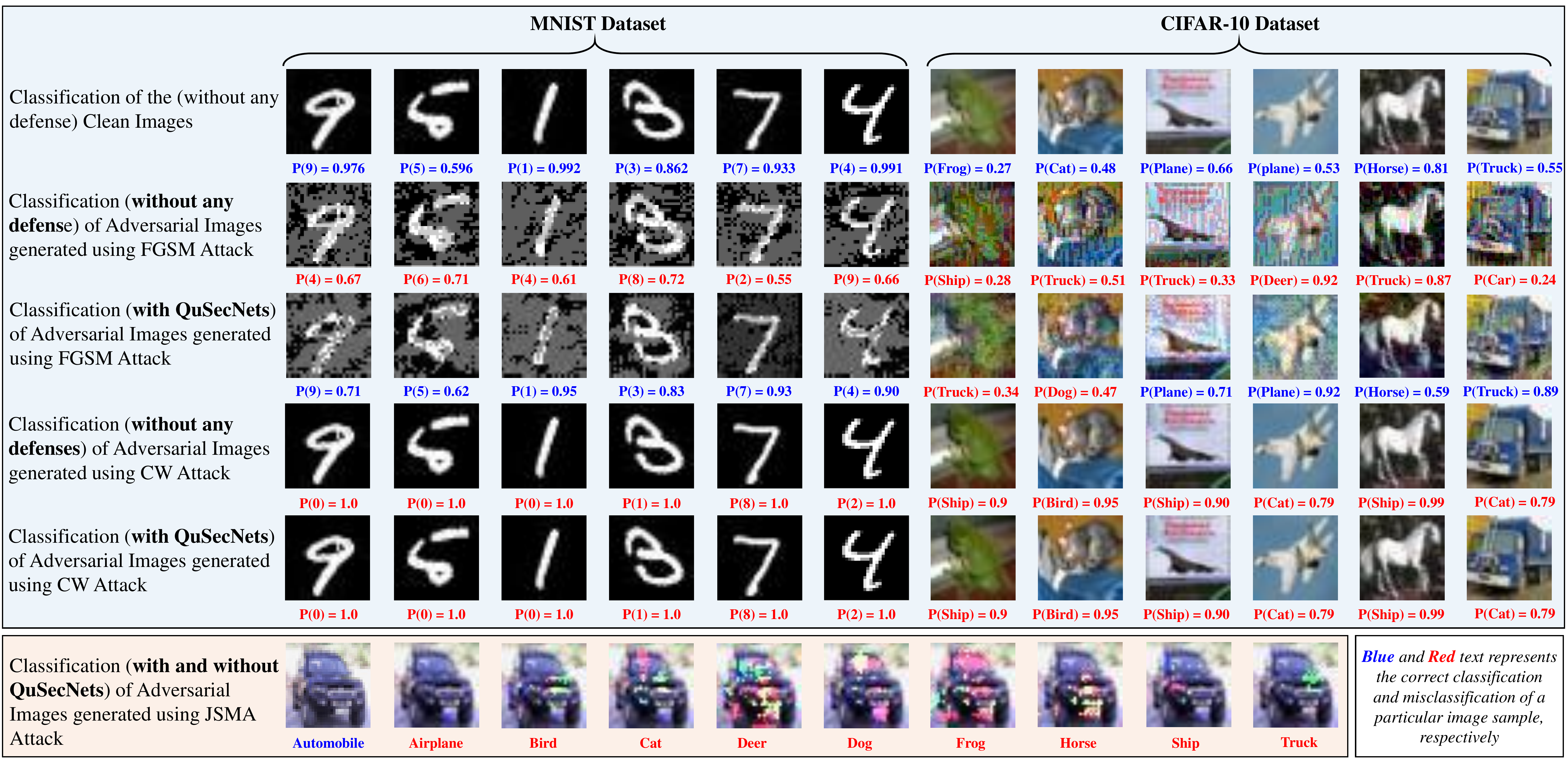}
    	\caption{\textit{Adversarial examples generated using the FGSM, the C\&W and the JSMA attacks from the images of the MNIST and the CIFAR-10 datasets to fool \textit{QuSecNets} along with the corresponding labels and probabilities. The results show that the perceptibility of adversarial noise is increased significantly by adding the proposed quantization layers (\textit{QuSecNets}). P(x) represents the probability with which the CNN classifies a particular input image as class ``x".}}
    	\label{fig:AE1}
    	\vskip -0.1in
    \end{figure*}
\subsection{Evaluation and Discussion}
Based on the experiments for the constant and the trainable quantization-based defense (\textit{QuSecNets}) against the state-of-the-art adversarial attacks, we make the following observations:

\begin{enumerate}[leftmargin=*]
    \item \textit{Impact of quantization on perceptibility (the visibility of adversarial noise in subjective and objective evaluation):} Fig.~\ref{fig:AE1} shows the effects of quantization on adversarial noise generated by the FGSM, the C\&W and the JSMA attacks. These results, especially in case of the JSMA attack, show that the adversarial noise is clearly visible in all of the generated adversarial examples (Fig.~\ref{fig:AE1}). In most of the cases, both constant and trainable quantization strategies neutralize the effects of adversarial noise. For example, the attack image of $9$ is mapped to label $4$ when \textit{QuSeNets} is not employed. However, when using our \textit{QuSeNets}, this attack is neutralized, and the quantized attack image is correctly classified though with less confidence ($P(9)=0.71$) than the clean image ($P(9)=0.976$). Based on these observations, we conclude that \textit{quantization, especially as a trainable layer in the CNN, can increase the robustness of a CNN against adversarial attacks by either neutralizing the effects of imperceptible attack noise or by making the attack noise perceptible for subjective and objective evaluations.}

    \item \textit{Increasing the number of quantization levels tends to decrease the classification accuracy under attack scenarios}, as shown in Fig.~\ref{fig:epsilon_trend}, because it reduces the interval between quantization levels which allows the attack noise to propagate. For example, consider $eplison = 0.2$, the classification accuracy decreases when the number of quantization levels is increased from $2$ to $6$. Moreover, an increase in the $epsilon$ results in further decrease in the classification accuracy. For example, the classification accuracy in case of $Q\_6$ and $eplison=0.3$ is less than the classification accuracy in case of $Q\_6$ and $eplison=0.2$. 
    
    \item In most of the cases, the trainable quantization in \textit{QuSecNets} performs significantly better than the constant quantization, as shown in Fig. \ref{fig:effect_VQ}, because by learning the quantization layer in the CNN, it makes the defense more robust against the attack noise. 
    
\end{enumerate} 

%% file: Sections/Sec5_Comparison.tex
\section{Comparison with State-of-the-Art Defenses } \label{comparison}
To demonstrate the effectiveness of the proposed defense, in this section, we present a comparison with the state-of-the-art defense mechanisms, i.e., the Feature Squeezing, the BRELU+GDA and the Dynamic Quantization Activation (DQA) \cite{DBLP:journals/corr/abs-1807-06714}.

\begin{figure}[!t]
	\centering
	\includegraphics[width=1\linewidth]{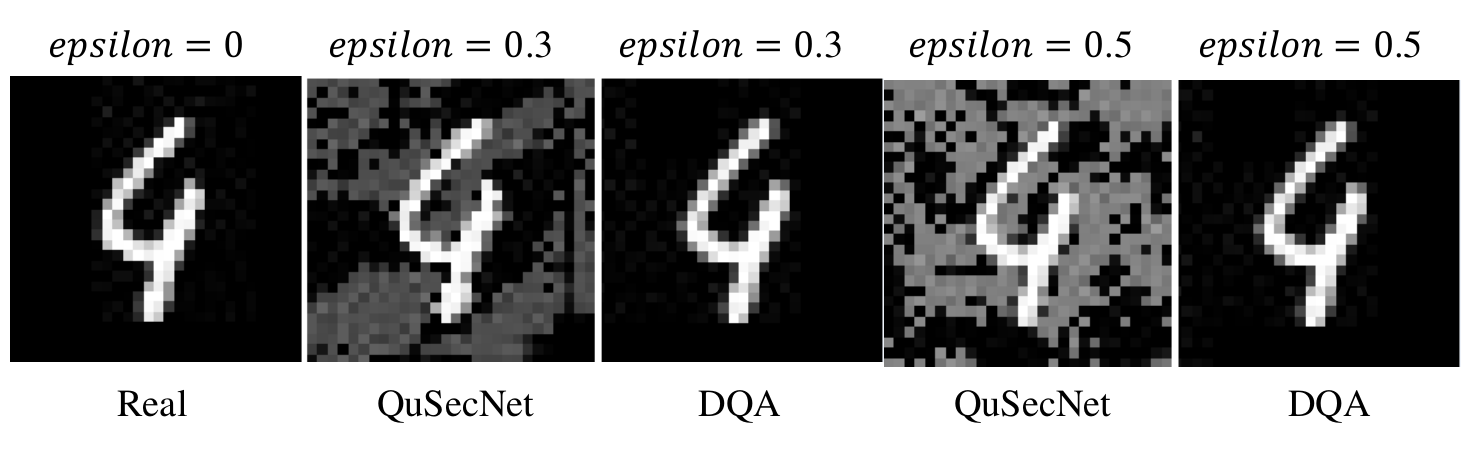} 
	\caption{\textit{Perceptibility comparison of adversarial examples generated using FGSM attack for QuSecNets and DQA \cite{DBLP:journals/corr/abs-1807-06714} with clean image.}}
	\vskip -0.1in
	\label{fig:comp_DQA}
\end{figure}

Experimental results show that the proposed trainable quantization increases the perceptibility of the adversarial noise in adversarial examples. For example, the attack noise generated by the FGSM attack is more perceptible in the case of TQ (\textit{QuSecNets}) as compared the case of DQA, as shown in Fig. \ref{fig:comp_DQA}.  

\begin{figure}[b]
	\centering
	\includegraphics[width=1\linewidth]{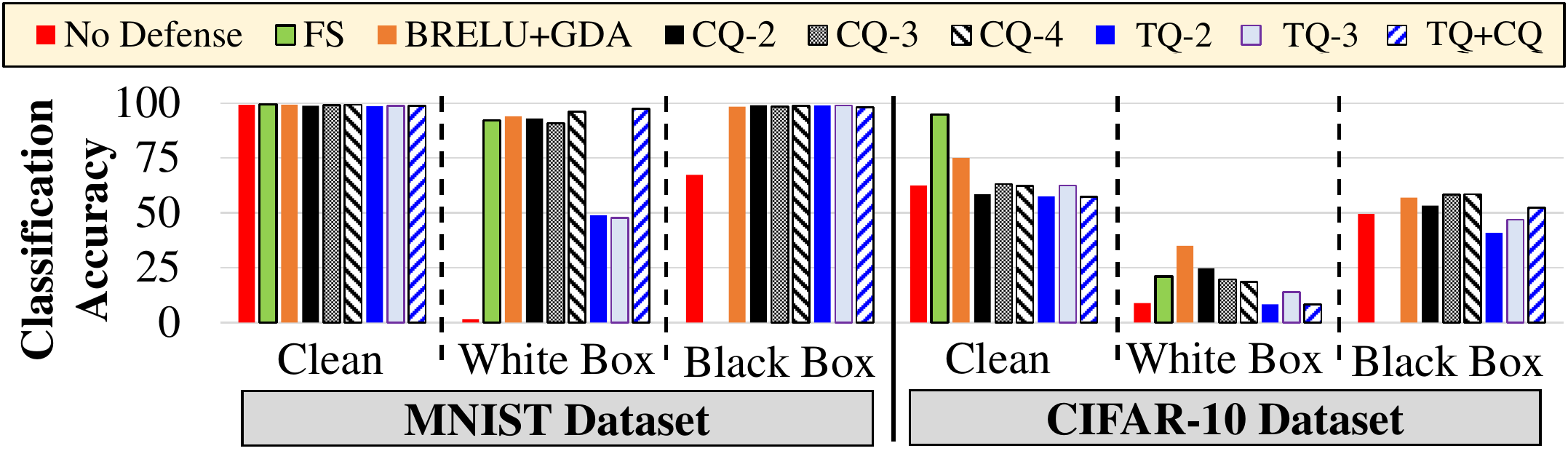}
	\caption{\textit{Comparison of the proposed strategies with the state-of-the-art defenses for FGSM attacks. Epsilon = 0.3 for the image from the MNIST and Epsilon = 0.1 for the images from CIFAR-10 datasets. Note, $n$ in CQ and TQ represents the number of quantization levels.}}
	\vskip -0.05in
	\label{fig:comp_FSGM}
\end{figure}
\textbf{Defenses against the FGSM attack:} Fig.~\ref{fig:comp_FSGM} reports our results for different defense mechanisms against the FGSM attack for the MNIST and the CIFAR-10 datasets. By analyzing these results, we make the following observations: 
\begin{enumerate}[leftmargin=*]
    \item The classification accuracy in attack-free scenario for the MNIST and CIFAR-10 datasets is slightly reduced when quantization is applied at the input.
    \item In the case of the white-box FGSM attack for the MNIST dataset, the constant quantization effectively improve the classification accuracy, i.e., from 1.48\% to 97.51\% which in case of the BReLU is 94\%. This is  because the quantization is inherently insensitive to small perturbations at the inputs. However, in case of the white-box FGSM attack for the CIFAR-10 dataset, \textit{QuSecNets} does not perform better because images in CIFAR-10 dataset are sensitive to quantization noise/error.
    \item In case of the black-box setting, the proposed \textit{QuSecNets} performs relatively better than the state-of-the-art defenses. This is because the CQ and TQ, in black-box setting, increase the minimum imperceptible noise required for misclassification. Note, here black-box setting corresponds to substitute model training followed by the FGSM attack.  
\end{enumerate}

\begin{figure}[!t]
	\centering
	\includegraphics[width=1\linewidth]{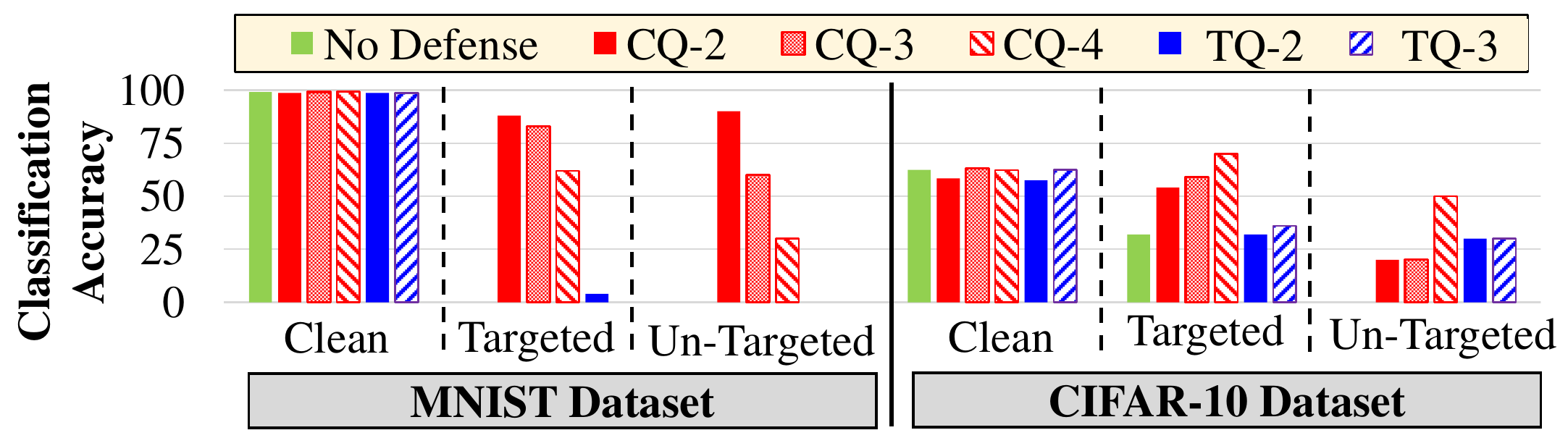}
	\caption{\textit{Comparison of the proposed defense strategies against the C\&W-L2 attack. Number of source samples for targeted C\&W-L2 attacks = 10, Number of iterations = 100, Maximum allowed perturbation in a pixel ($epsilon$) = 0.1. Note, $n$ in CQ and TQ represents the number of quantization levels.}}
	\vskip -0.1in
	\label{fig:comp_CW}
\end{figure}
\textbf{Defenses against the C\&W-L2 attack:} Fig.~\ref{fig:comp_CW} shows the comparison of the proposed defense strategies, i.e, CQ and TQ with different number of quantization levels, against the targeted and the un-targeted C\&W-L2 attacks. 
The figure shows that \textit{QuSecNets} results in significant improvement in the robustness of CNNs for both the MNIST and CIFAR-10 data sets, when compared to the CNNs with no defense mechanism (i.e., classification accuracy under C\&W-L2 attack is 0\%). 
When employing QuSecNets, the classification accuracy for the MNIST data set under C\&W-L2 attack is 88\% (54\% for CIFAR-10) and 90\% (50\% for CIFAR-10) for the white-box and black-box settings, respectively.
The proposed defense methodology (\textit{QuSecNets}) is not robust against the C\&W attacks if the maximum allowed perturbation in a single pixel (i.e, $epsilon$) is relatively large. For example, if $epsilon$ is increased from 0.1 to 0.2, the classification accuracy drops significantly. 

\begin{figure}[!t]
	\centering
	\includegraphics[width=1\linewidth]{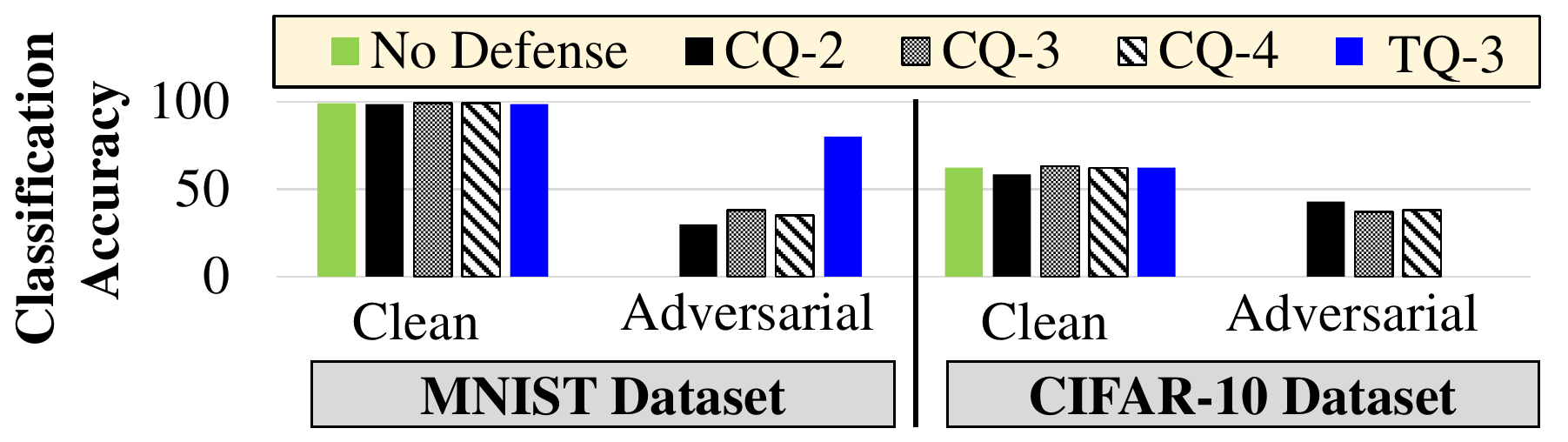}
	\caption{\textit{Comparison of the proposed quantization-based defenses (\textit{QuSecNets}) against the targeted JSMA attack. Number of source samples = 10, Number of iterations = 100. Note, $n$ in CQ and TQ represents the number of quantization levels.}}
	\vskip -0.05in
	\label{fig:comp_JSMA}
\end{figure}
\textbf{Defenses against the targeted JSMA attack:} Fig.~\ref{fig:comp_JSMA} reports the results of \textit{QuSecNets} against the targeted JSMA attack. The \textit{QuSecNets} does not effectively counter the JSMA attack because this attack adds high intensity noise in a few pixels of the input image, instead of a small distributed perturbation. However, the classification accuracy for such attacks can be significantly improved by other strategies such as median filters \cite{DBLP:conf/ndss/Xu0Q18} or input drop-out. 

In general, we observe that quantization-based defense against adversarial attacks shows better results in terms of classification accuracy for images from the MNIST dataset as compared to the images from the CIFAR-10 dataset (See Figure~\ref{fig:comp_FSGM} to Figure~\ref{fig:comp_JSMA}) because of the clustered distribution of the MNIST dataset. 

%% file: Sections/Sec6_Conclusion.tex
\section{Conclusion}\label{conclusion}
In this paper, we proposes to leverage the insensitivity towards small perturbation and dynamic nature of trainable quantization for developing a defense mechanism, \textit{QuSecNets}, against adversarial attacks. This methodology introduces an additional layer at the input of DNNs, to increase the perceptibly of adversarial noise in the adversarial examples, and thereby make them easily detectable by subjective and objective evaluation. To demonstrate the effectiveness of the proposed methodology, we evaluated our approach against some of the state-of-the-art adversarial attacks (i.e., FGSM, C\&W and JSMA) and compared it with the corresponding state-of-the-art defenses. We empirically demonstrated that integrating the Trainable Quantization Layer can significantly hardens the DNNs. 

%% file: Sections/Sec_ack.tex
\section*{Acknowledgement}
This work was partially supported by the Erasmus+ International Credit Mobility (KA107). 